\documentclass[conference]{IEEEtran}
\usepackage{times}
\usepackage{multirow}
\usepackage{amsmath}
\usepackage{xcolor}

\usepackage{graphicx}
\usepackage{capt-of} 
\usepackage{amsfonts}
\usepackage{amssymb}
\usepackage[ruled,vlined,linesnumbered]{algorithm2e}
\usepackage{amsmath}
\usepackage{booktabs, tabularx, array}
\usepackage{multirow}
\usepackage{balance}

\usepackage[numbers]{natbib}
\usepackage{multicol}
\usepackage[bookmarks=true]{hyperref}

\renewcommand{\tablename}{Table}
\makeatletter
\let\@IEEEorig@makecaption\@makecaption
\long\def\@makecaption#1#2{%
  \ifx\@captype\@IEEEtablestring%
    \@IEEEtablecaptionsepspace
    \setbox\@tempboxa\hbox{\normalfont\footnotesize {#1.}~~ #2}%
    \ifdim \wd\@tempboxa >\hsize%
      \setbox\@tempboxa\hbox{\normalfont\footnotesize {#1.}~~ }%
      \parbox[t]{\hsize}{\normalfont\footnotesize\noindent\unhbox\@tempboxa#2}%
    \else%
      \ifCLASSOPTIONconference
        \hbox to\hsize{\normalfont\footnotesize\hfil\box\@tempboxa\hfil}%
      \else
        \hbox to\hsize{\normalfont\footnotesize\box\@tempboxa\hfil}%
      \fi
    \fi
  \else
    \@IEEEorig@makecaption{#1}{#2}%
  \fi}
\newcommand{\onedot}{\ifx\@let@token.\else.\null\fi}
\makeatother

\newcommand{\action}{\ensuremath{\mathbf{a}}}
\newcommand{\obs}{\ensuremath{o}}
\newcommand{\dataset}{\ensuremath{\mathcal{D}}}

\makeatother

\begin{document}

% paper title
\title{VLS: Steering Pretrained Robot Policies via Vision–Language Models}

% You will get a Paper-ID when submitting a pdf file to the conference system
% \author{Author Names Omitted for Anonymous Review. Paper-ID [8]}
\author{
    \IEEEauthorblockN{
        Shuo Liu$^{1,4}$,
        Ishneet Sukhvinder Singh$^{2}$,
        Yiqing Xu$^{3,4}$,
        Jiafei Duan$^{1,4,*}$,
        Ranjay Krishna$^{1,4,*}$
    }
    \IEEEauthorblockA{$^{1}$University of Washington \quad $^{2}$University of Oxford}
    \IEEEauthorblockA{$^{3}$National University of Singapore \quad $^{4}$Allen Institute for Artificial Intelligence}
    \IEEEauthorblockA{$^{*}$Co-advising
}
    \IEEEauthorblockA{\href{https://vision-language-steering.github.io/webpage/}{\textcolor{blue}{vision-language-steering.github.io}}}
}

%\author{\authorblockN{Michael Shell}
%\authorblockA{School of Electrical and\\Computer Engineering\\
%Georgia Institute of Technology\\
%Atlanta, Georgia 30332--0250\\
%Email: mshell@ece.gatech.edu}
%\and
%\authorblockN{Homer Simpson}
%\authorblockA{Twentieth Century Fox\\
%Springfield, USA\\
%Email: homer@thesimpsons.com}
%\and
%\authorblockN{James Kirk\\ and Montgomery Scott}
%\authorblockA{Starfleet Academy\\
%San Francisco, California 96678-2391\\
%Telephone: (800) 555--1212\\
%Fax: (888) 555--1212}}

% avoiding spaces at the end of the author lines is not a problem with
% conference papers because we don't use \thanks or \IEEEmembership

% for over three affiliations, or if they all won't fit within the width
% of the page, use this alternative format:
% 
%\author{\authorblockN{Michael Shell\authorrefmark{1},
%Homer Simpson\authorrefmark{2},
%James Kirk\authorrefmark{3}, 
%Montgomery Scott\authorrefmark{3} and
%Eldon Tyrell\authorrefmark{4}}
%\authorblockA{\authorrefmark{1}School of Electrical and Computer Engineering\\
%Georgia Institute of Technology,
%Atlanta, Georgia 30332--0250\\ Email: mshell@ece.gatech.edu}
%\authorblockA{\authorrefmark{2}Twentieth Century Fox, Springfield, USA\\
%Email: homer@thesimpsons.com}
%\authorblockA{\authorrefmark{3}Starfleet Academy, San Francisco, California 96678-2391\\
%Telephone: (800) 555--1212, Fax: (888) 555--1212}
%\authorblockA{\authorrefmark{4}Tyrell Inc., 123 Replicant Street, Los Angeles, California 90210--4321}}

\makeatletter
\let\@oldmaketitle\@maketitle%
\renewcommand{\@maketitle}{\@oldmaketitle%
\setcounter{figure}{0} %
\centering
\includegraphics[width=1.0\textwidth]{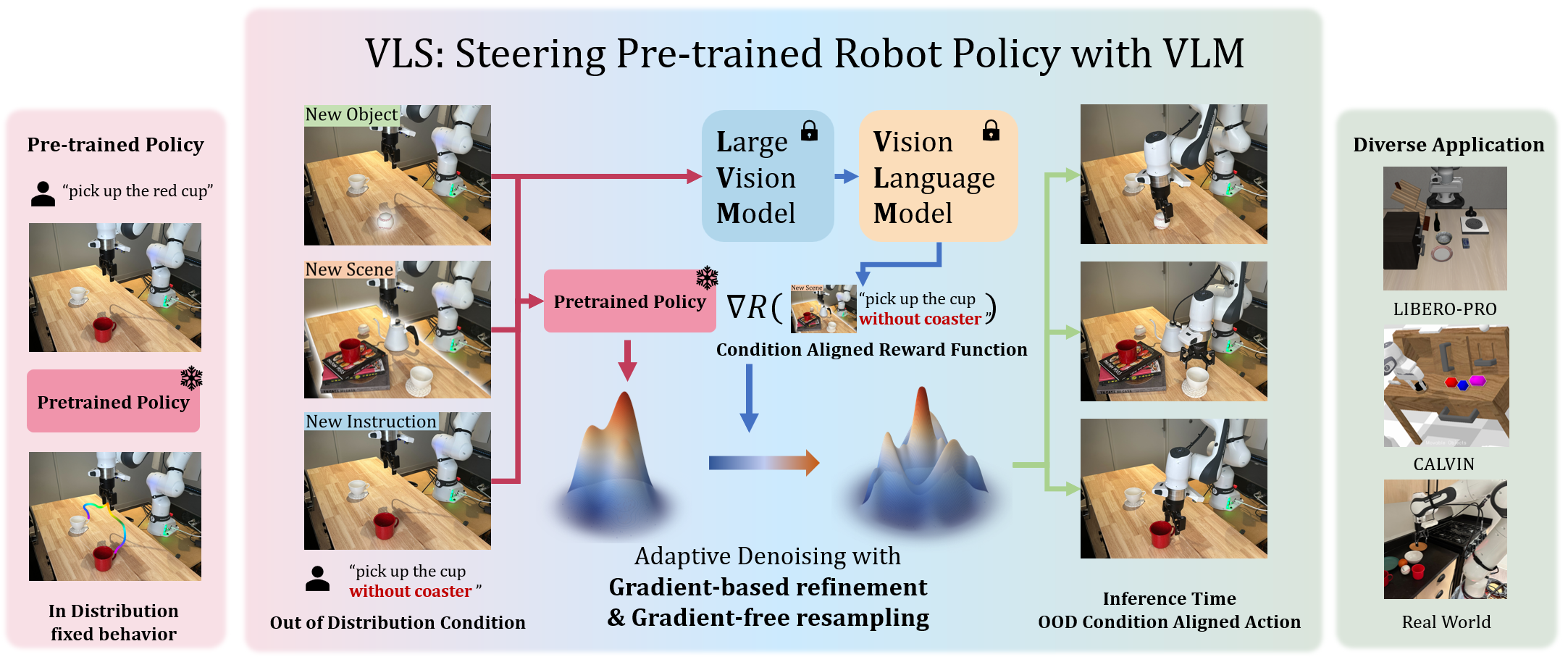}
\captionof{figure}{We present \textbf{Vision--Language Steering (VLS)}, a training-free framework for inference-time steering of frozen generative robot policies. Our core idea is to leverage the open-world understanding capabilities of VLMs to generate reward functions for partially denoised action proposals, helping the base policy successfully operate in out-of-distribution (OOD) scenarios such as object changes, scene changes or instruction changes by correcting the denoising path. VLS demonstrates excellent performance in simulation benchmarks as well as real-world experiments, proving its effectiveness.}
\label{fig:front-fig}
\vspace{-0.3in}
\bigskip}
\makeatother

\maketitle

\begin{abstract}
Why do pretrained diffusion or flow-matching policies fail when the same task is performed near an obstacle, on a shifted support surface, or amid mild clutter? Such failures rarely reflect missing motor skills; instead, they expose a limitation of imitation learning under train–test shifts, where action generation is tightly coupled to training-specific spatial configurations and task specifications. Retraining or fine-tuning to address these failures is costly and conceptually misaligned, as the required behaviors already exist but cannot be selectively adapted at test time. We propose Vision–Language Steering (VLS), a training-free framework for inference-time adaptation of frozen generative robot policies. VLS treats adaptation as an inference-time control problem, steering the sampling process of a pretrained diffusion or flow-matching policy in response to out-of-distribution observation–language inputs without modifying policy parameters. By leveraging vision–language models to synthesize trajectory-differentiable reward functions, VLS guides denoising toward action trajectories that satisfy test-time spatial and task requirements. Across simulation and real-world evaluations, VLS consistently outperforms prior steering methods, achieving a 31\% improvement on CALVIN and a 13\% gain on LIBERO-PRO. Real-world deployment on a Franka robot further demonstrates robust inference-time adaptation under test-time spatial and semantic shifts.

\end{abstract}

\IEEEpeerreviewmaketitle

\section{Introduction}

Once a child learns to place a cup at the center of a table, they have not merely mastered a single task.
The same motor skill generalizes to placing the cup near an edge, atop a stack of books, or inside a crowded cabinet. For humans, motor execution naturally transfers across such spatial and task variations \cite{duan2021spacesimulatorphysicalinteractions,duan2022survey}. For robots, however, skill execution is often tightly coupled to the specific spatial configurations and instructions encountered during training. As a result, even state-of-the-art manipulation policies ~\cite{khazatsky2024droid,o2024open,walke2023bridgedata,chi2023diffusionpolicy,barreiros2025careful,rt12022arxiv,brohan2023rt2,reed2022gato,driess2023palme,ahn2022saycan,jiang2022vima} can fail when the observation or instruction at test time deviates from the training distribution: a robot that succeeds at placing an object at the center of a table may hesitate, collide, or miss entirely when asked to place it near an edge. These failures do not reflect missing motor capability, but rather the absence of a mechanism to adapt existing skills to new spatial requirements at test time.

Recent advances in robot learning have produced expressive pretrained policies, particularly those based on diffusion or flow-matching objectives, that achieve strong in-distribution performance~\cite{black2024pi0,lipman2022flowmatching,yu2025compose}.
However, these generative policies remain brittle under out-of-distribution (OOD) observation--language inputs \cite{pumacay2024colosseum}, where the required motor behaviors are already present in the training data but must be executed under altered spatial structure.
Addressing such failures through retraining or fine-tuning is costly and conceptually misaligned, as it attempts to relearn behaviors rather than control their execution~\cite{yuan2024policydecorator,zhu2025usr,wagenmaker2025dsrl}.
Expanding the training distribution to cover all possible spatial variations is therefore a brute-force solution to what is fundamentally an inference-time control problem.

In this work, we propose \textbf{Vision--Language Steering (VLS)}, a training-free framework for inference-time adaptation of pretrained robotic policies.
VLS operates on a frozen base policy and addresses OOD inputs—joint observation–language pairs \((o,l)_{\textit{OOD}}\) that lie outside the expert dataset—by steering the policy’s sampling process at test time.
Rather than modifying policy parameters, VLS reshapes the action distribution during inference so that generated trajectories satisfy the spatial and task structure implied by \((o,l)_{\textit{OOD}}\).
This formulation explicitly decouples skill execution from OOD task specification:
the base policy provides reusable motor primitives, while inference-time steering controls how those primitives are composed and instantiated under OOD inputs.

Our approach is inspired by inference-time steering techniques developed for large language models and image generation models ~\cite{dathathri2019pplm,dhariwal2021guided,ho2022cfg,meng2021sdedit,kim2022diffusionclip,tumanyan2023pnpdiffusionfeatures}, where a pretrained model’s output distribution is reshaped to elicit desired behaviors without additional training. VLS extends this paradigm to robotics by treating action generation as a controllable denoising process. Specifically, we leverage vision--language models (VLMs) to interpret OOD observation–language inputs, decompose tasks into execution stages, and synthesize differentiable reward functions that score action proposals with respect to spatial structure.
These rewards provide dense, trajectory-level gradients that guide diffusion or flow-matching policies during inference. By grounding OOD inputs into geometric representations and injecting reward-based guidance into the denoising process, VLS enables existing skills to be executed reliably under spatial variation and unseen task specifications, while preserving the robustness of the frozen base policy.

We evaluate VLS in both simulation and real-world settings under observation and language shifts at test time. In simulation, we benchmark on CALVIN~\cite{calvin} and LIBERO-PRO~\cite{zhou2025liberopro}, two widely used manipulation suites that explicitly stress inference-time adaptation to out-of-distribution observation–language inputs. On CALVIN, VLS consistently outperforms prior inference-time steering methods such as ITPS~\cite{wang2024itps} and DynaGuide~\cite{du2025dynaguide}, achieving up to a $31\%$ absolute improvement in success rate on long-horizon tasks. On LIBERO-PRO, VLS improves the success rate of frozen VLA policies, including OpenVLA~\cite{kim24openvla}, $\pi_0$~\cite{black2024pi0}, variants of $\pi_{0.5}$ ~\cite{black2025pi05, lerobot_pi05} by up to 13\% under both spatial (object layout) and semantic (task specification) perturbations. Finally, real-world experiments on a Franka robot demonstrate that VLS enables stable execution of multi-stage, language-specified tasks under unseen object appearances, positional changes, and target substitutions, validating its effectiveness for practical deployment.

\section{Related Work}

\subsection{Imitation-Trained Policies under Small Environment Shifts}

Large-scale imitation learning has enabled impressive generalist and VLM-conditioned robot policies, including diffusion- and flow-matching generators \cite{barreiros2025careful, khazatsky2024droid, o2024open,  walke2023bridgedata}. However, a consistent failure mode remains: even mild changes in scene geometry or context (e.g., clutter, support-surface shifts, different object layouts) can cause sharp degradation. This brittleness is a known limitation of imitation learning: policies learn correlations between action and training context, and thus do not reliably produce constraint-satisfying behaviors when the environment configuration departs from the training manifold. This motivates methods that adapt execution at test time without assuming new data coverage.

\subsection{VLM-based Scene Understanding with Re-optimization}

A common way is to use VLMs to generate scene representations that improves spatial understanding and then re-optimize actions online via planning, search, or iterative refinement \cite{huang2023voxposer, huang2024rekep, kumar2024open}. While these approaches can handle unseen observations, they typically require rollouts, repeated evaluation, or online optimization loops, which are computationally heavy and often incompatible with real-time control. Moreover, they shift the burden of generalization to optimization at deployment, whereas our goal is to retain the pretrained policy as the skill prior and adapt behavior through lightweight inference-time control.

\subsection{Inference-time Steering of Generative Policies}
Most related to our approach are inference-time methods that steer the sampling of a pre-trained policy.

\textbf{Value/critic-guided steering.} V-GPS re-ranks actions using an offline-learned value function to improve generalist policies without updating the backbone~\cite{nakamoto2024vgps}. For diffusion policies, VGD injects gradients from a learned value/Q model into denoising to bias trajectories toward higher value while keeping the policy frozen~\cite{ye2025vgd}. These methods provide dense guidance, but they do so through an auxiliary learned objective, which can effectively reshape the policy toward the critic’s preferences. However, we view this as undesirable as the base policy should remain the invariant, and only the test-time constraints should modulate execution.

\textbf{Dynamics/world-model guided steering.} DynaGuide uses an external dynamics model to guide denoising, enabling multi-objective steering while preserving the diffusion prior~\cite{du2025dynaguide}. Latent Policy Barrier uses a learned dynamics model to predict and optimize future latent states so trajectories remain within an expert manifold under covariate shift~\cite{sun2025lpb}. These approaches increase dependence on predictive modeling and rollout-style evaluation, and can become sensitive to model error and inference cost as it pushes adaptation burden into heavier test-time optimization.

\textbf{Human/VLM-in-the-loop steering and verification.} ITPS steers generative sampling through human interaction signals at inference time~\cite{wang2024itps}. FOREWARN uses VLMs as open-vocabulary verifiers to select among candidate plans~\cite{wu2025forewarn}, and Do What You Say similarly checks reasoning–action faithfulness by filtering candidate action sequences using VLM-based alignment~\cite{wu2025dowhatyousay}. These methods demonstrate the power of semantic feedback, but their supervision is typically discrete and sparse, which forces adaptation to occur through selection/rejection over candidates rather than through continuous, differentiable steering within generation, making them sample-inefficient when the desired behavior requires fine-grained constraint satisfaction. The work most closely related to VLS is VLA-Pilot~\cite{li2025towards}, but our focus is on guiding pre-trained policies to handle OOD scenarios, combining gradient-guided denoising processes with dynamic stage transitions, and conducting extensive testing in both simulation and real-world.

\textbf{Online improvement without finetuning the base policy.} Policy Decorator adds a residual refinement policy for online correction while preserving the backbone imitation policy~\cite{yuan2024policydecorator}. USR unifies latent steering with residual refinement via a lightweight actor for online improvement of diffusion policies~\cite{zhu2025usr}. DSRL optimizes in diffusion latent/noise space to enable fast online improvement with black-box access to the base policy~\cite{wagenmaker2025dsrl}. These methods rely on online learning or interaction, whereas we focus on pure inference-time adaptation with no additional training at deployment.

\section{Problem Formulation}
\subsection{The OOD Dilemma in Imitation Learning}

Imitation learning aims to learn a policy $\pi_{\theta}$ from an expert demonstration dataset $\dataset_{\textit{expert}} = \{(\obs_i, \action_i), l_i\}_{i=1}^N$, modeling the state-conditional action distribution $p(\action|\obs)$. Typically, at environment time step $t$, the training target of a policy $\pi_{\theta}$ is to maximize the likelihood of an action chunk $\mathbf{a}_{t:t+T}$ with chunk horizon $T$, conditioned on the observation $o$ (typically RGB images and robot proprioception) and language instruction $l$ :
\begin{equation}\label{eq:train_target}
 \max_{\theta} \mathbb{E}_{(\mathbf{a}, o, l) \sim \mathcal{D}_{\textit{expert}}} \left[ \sum_{t=1}^T \log \pi_{\theta} (\mathbf{a}_{t:t+T} | o_t, l) \right].
\end{equation}
After training, the policy $\pi_{\theta}$'s weight $\theta$ can be frozen, which we refer to as the \textbf{base policy} $\pi^\star$. However, this training objective is inherently static and distribution-dependent. When the policy is deployed in real world, it inevitably encounters out-of-distribution (OOD) scenarios $\{o, l\}_{\textit{OOD}} \notin \dataset_{expert}$, which ranging from observation shift ($o_{\textit{OOD}}$) such as change of visual backgrounds or object layouts to semantic ambiguity ($l_{\textit{OOD}}$) such as unseen instructions. 
Since the base policy $\pi^\star$ tends to overfit on the spatial and semantic correlations present in the training manifold, it exhibits severe brittleness when faced with such OOD scenarios~\cite{ross2011dagger}. 

\subsection{Diffusion and Flow Matching Policies}

In recent years, denoising generative models have become a cornerstone for imitation learning~\cite{chi2023diffusionpolicy}. These models learn to transform a simple Gaussian distribution into a complex target action distribution $q(\action_{t:t+T}^0)$ through forward diffusion and reverse denoising~\cite{ho_ddpm_2020}. The forward process gradually adds Gaussian noise to the clean action trajectory $\action_{t:t+T}^0$, transforming it into a Gaussian distribution $\mathbf{a}_{t:t+T}^K \sim \mathcal{N}(\mathbf{0}, \mathbf{I})$. A network $\epsilon(\mathbf{a}^k,o,l,k)$ is then trained to predicted the added noise at denoising step $k \in [0, 1, ..., K]$, conditioned on the observation $o$ and instruction $l$. The reverse process samples action from $\mathbf{a}_{t:t+T}^K \sim \mathcal{N}(\mathbf{0}, \mathbf{I})$ and applies the update

\begin{equation}\label{eq:diffusion_transition}
\mathbf{a}_{t:t+T}^{k-1} = \frac{1}{\sqrt{\alpha_k}} \left( \mathbf{a}_{t:t+T}^k - \frac{1-\alpha_k}{\sqrt{1-\bar{\alpha}_k}} \epsilon(\mathbf{a}_{t:t+T}^k, o, l, k) \right) + \sigma_k \mathbf{z},
\end{equation}
where $\mathbf{z} \sim \mathcal{N}(\mathbf{0}, \mathbf{I})$, $\alpha_k$ and $\bar{\alpha}_k$ are noise schedule coefficients, eventually producing $\mathbf{a}_{t:t+T}^0$ that approximates $q(\mathbf{a}_{t:t+T}^0)$. 
% $\sigma_k = 0$ for DDPM sampling [DDPM] and when $\sigma_k = 0$ for DDIM sampling [DDIM].

Beyond denoising diffusion, flow matching simplifies the denoising process by learning a continuous velocity field $v$. While flow-matching also iteratively refines samples from a Gaussian distribution, different from discrete indices $\{K, K-1, \dots, 0\}$ that often used in diffusion models, flow matching utilizes a continuous time interval $[0, 1]$, where $k=1$ and $k=0$ correspond to the noise distribution and the clean action trajectory, respectively. Since both variables fundamentally represent the progression of the denoising process, we unify the notation by using $k$ for both frameworks to avoid redundant symbolic definitions and potential ambiguity. Flow matching models the transition of distribution as an Ordinary Differential Equation (ODE):

\begin{equation}\label{eq:flow_transition}
\frac{d\action_{t:t+T}^k}{dt} = v(\action_{t:t+T}^k,o, l, k).
\end{equation}

\subsection{Problem Formulation}
Given a pre-trained \textbf{base policy} $\pi^\star$, our goal is to enable it to adapt to OOD scenarios $\{o, l\}_{\textit{OOD}} \notin \dataset_{expert}$ at inference time without fine-tuning. To enable $\pi^\star$ adapt to the new condition $(o, l)_{\textit{OOD}}$, we leverage Classifier Guidance~\cite{dhariwal2021guided} to steer the sampling process of the base policy. The core idea is to find a guidance function and use the gradient $g = \nabla_{\action_{t:t+T}^k} \log p((o, l)_{\textit{OOD}} | \action_{t:t+T}^k)
$ which represent the score of joint distribution of action proposal $\action_{t:t+T}^k$ and OOD condition $(o, l)_{\textit{OOD}}$, to guide the direction of denoising. We provide a detailed derivation of gradient-based steering for diffusion and flow-matching policies in Appendix.

% \yiqing{TODO: $y = (\obs_{OOD}, \l_{OOD})$, let's stick to one of them. if the guidance function takes in ($\obs_{OOD}, \l_{OOD}$), then it is not about context-aware constraints, but new conditions that are adjacent to the expert data used to train the fm/dp policy. We need to find a more accurate description here. }

For diffusion models, the modified the noise prediction is:
\begin{equation}\label{eq:diffusion_guidance}
    \begin{aligned}
    \hat{\epsilon}
    % &= \epsilon(\mathbf{a}_{t:t+T}^k, (o,l)_{\textit{OOD}}, k)
    %    - \lambda \cdot \sqrt{1-\bar{\alpha}_k}
    %      \nabla_{\mathbf{a}_{t:t+T}^k} \log p(y|\mathbf{a}_{t:t+T}^k) \\
    &= \epsilon(\action_{t:t+T}^k, (o, l)_{\textit{OOD}}, k)
       - \lambda \cdot \sqrt{1-\bar{\alpha}_k} \, \cdot g(\action_{t:t+T}^k, (o, l)_{\textit{OOD}}),
    \end{aligned}
\end{equation}
where $\lambda$ is the guidance scale hypermeter to control the guidance strength. Similar to diffusion model, a flow matching policy can be steered to accommodate the condition $y = (o, l)_{\textit{OOD}}$ by controlling the predicted velocity field:

\begin{equation}\label{eq:flow_guidance}
\begin{aligned}
\hat{v} 
% &= v(\action_{t:t+T}^k,o,k) + \lambda \nabla_{\mathbf{a}_{t:t+T}^k} \log p(y|\mathbf{a}_{t:t+T}^k) \\
&= v(\action_{t:t+T}^k, (o, l)_{\textit{OOD}}, k) + \lambda \cdot g(\action_{t:t+T}^k, (o, l)_{\textit{OOD}}).
\end{aligned}
\end{equation}

The main challenge lies in modeling the gradient guidance function \(g(\cdot)\).
In real-world OOD deployment, the conditioning input \((o, l)_{\textit{OOD}}\) is not a simple class label, but a structured specification that implicitly encodes spatial and semantic constraints.
A valid guidance function must therefore flexibly and accurately model
\(\log p\!\left((o, l)_{\textit{OOD}} \mid \action_{t:t+T}^k\right)\),
while satisfying two requirements:
(1) it must correctly interpret the geometry and logical structure induced by the OOD condition, and
(2) it must provide dense, informative gradients with respect to the proposed action trajectory.

\section{Our Approach: VLS}

\begin{figure*}[t]
    \centering
    \includegraphics[width=\textwidth]{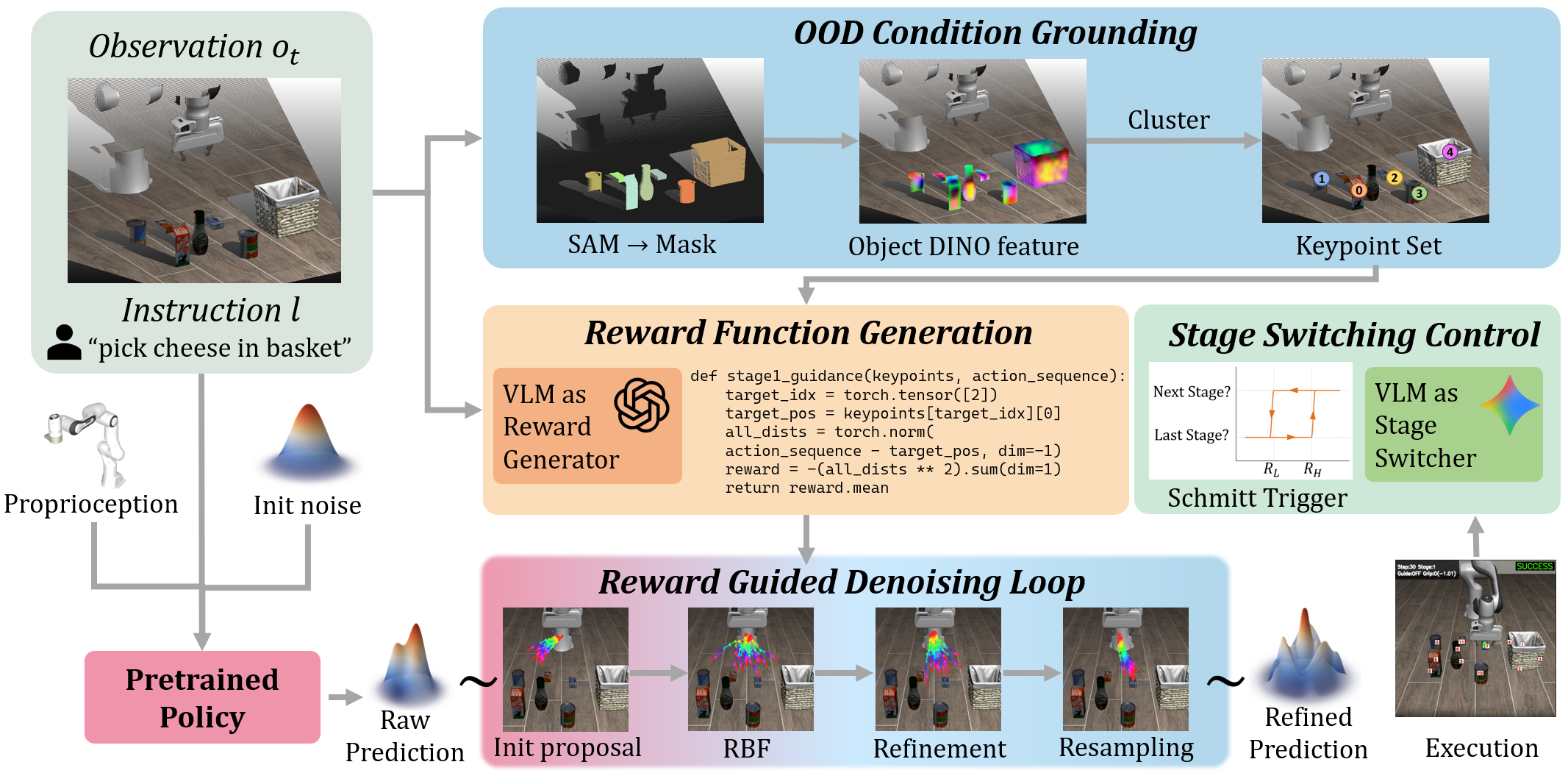}
    \caption{\textbf{VLS pipeline overview.} At environment time step $t$, given RGB-D observation $o_t$ and language instruction $l$, VLS firstly utilize the Segment Anything Model (SAM~\cite{sam}) and DINOv2~\cite{dino} feature to ground condition into a set of spatial keypoints $\mathcal{P}$. Subsequently, a Vision-Language Model will be queried to generates a series of stage-aware differentiable programmatic reward functions $\{\mathcal{R}_s\}_{s=1}^S$, based on observation, task instruction and keypoints, which are used to guide the action generation process of the frozen base policy $\pi^\star$: during the denoising sampling loop, the system precisely corrects action trajectories by injecting reward gradients, incorporating RBF~\cite{tdp} repulsion terms and a Feynman–Kac~\cite{singhal2025general} based resampling mechanism to rapidly converge to high-reward regions while maintaining sampling diversity. Finally, VLS constructs a closed-loop stage switching system based on reward feedback, utilizing adaptive guidance strength and Schmitt-trigger~\cite{schmitt1938thermionic} switching logic to monitor execution progress, thereby automatically triggering phase transitions or retry strategies when facing physical uncertainties (such as object displacement or manipulation failures), ensuring robust completion of long-horizon manipulation tasks in OOD environments.}
    \label{fig:pipeline}
\end{figure*}

\label{sec:method}

The core of Vision--Language Steering (VLS) is to approximate the guidance signal
\(
g \triangleq \nabla_{\action_{t:t+T}^k}\log p\!\left((o,l)_{\textit{OOD}} \mid \action_{t:t+T}^k\right)
\)
for inference-time adaptation under OOD inputs.
Since the likelihood term is not directly available in real-world deployment, VLS constructs a differentiable surrogate score
\(\mathcal{R}\) that maps an action proposal to how well it satisfies the spatial and semantic constraints implied by the OOD input pair \((o,l)_{\textit{OOD}}\):
\begin{equation}\label{eq:likelihood_estimatation}
\mathcal{R}(\action_{t:t+T}^k, (o, l)_{\textit{OOD}}) \approx \log p\!\left((o, l)_{\textit{OOD}} \mid \action_{t:t+T}^k\right).
\end{equation}
By design, \(\mathcal{R}\) is differentiable with respect to \(\action_{t:t+T}^k\), enabling gradient guidance
\(g \approx \nabla_{\action_{t:t+T}^k}\mathcal{R}(\action_{t:t+T}^k,(o,l)_{\textit{OOD}})\)
to steer the denoising trajectory of the frozen base policy \(\pi^\star\) without fine-tuning.
As illustrated in Figure \ref{fig:pipeline}, VLS instantiates this pipeline with three components:

\begin{enumerate}
    \item \textbf{OOD input grounding and VLM-generated differentiable scoring.} Ground the OOD input pair \((o,l)_{\textit{OOD}}\) into a compact geometric scaffold \(\mathcal{P}\) of task-relevant 3D keypoints, and use a vision--language model to synthesize stage-wise, programmatic reward functions \(\mathcal{R}(\cdot,\mathcal{P})\) that provide differentiable scores over action proposals (Sec.~\ref{subsection 4.1}).
    \item \textbf{Inference-time denoising guidance.} Inject \(\nabla_{\action}\mathcal{R}_s\) into the diffusion or flow-matching denoising updates, combining gradient-based refinement with particle-level diversity and resampling to steer the action distribution under OOD inputs (Sec.~\ref{subsection 4.2}).
    \item \textbf{Closed-loop execution control and stage switching.} Use execution feedback to adaptively regulate the guidance strength across action chunks and to determine transitions between stage-specific reward functions, enabling stable multi-stage execution under physical uncertainty (Sec.~\ref{subsection 4.3}).
\end{enumerate}

Full implementation details corresponding to each component of VLS, including the VLM prompt design and reward function structure, are provided in Appendix.

\subsection{OOD Input Grounding and Reward Generation}
\label{subsection 4.1}

To construct the surrogate score
\(\mathcal{R}(\action_{t:t+T}^k,(o,l)_{\textit{OOD}})\),
VLS must map the high-dimensional OOD input pair \((o,l)_{\textit{OOD}}\)
to a differentiable function over the action space.
We achieve this by (i) grounding the OOD input into a compact set of task-relevant spatial variables,
and (ii) synthesizing programmatic, differentiable reward functions over these variables.

\subsubsection{OOD Input Grounding}

Given an OOD input pair \((o,l)_{\textit{OOD}}\),
VLS first identifies the objects and regions relevant to the manipulation task using a vision--language model (VLM).
For each identified object, we apply the Segment Anything Model (SAM~\cite{sam}) to obtain a set of object masks
\(\mathcal{M}\).
Following~\cite{huang2024rekep}, we extract semantically aligned dense visual features using DINOv2~\cite{dino},
producing a patch-wise feature map
\(\Phi \in \mathbb{R}^{H \times W \times d}\),
which is filtered using the corresponding object masks.

To recover physical structure, masked pixels are reprojected into a 3D point cloud using depth information.
Each point is represented by the concatenation of its DINO feature (\(d\) dimensions)
and its 3D spatial coordinates, yielding a \((d+3)\)-dimensional representation.
We cluster these object-centric point clouds to obtain a set of task-relevant keypoints
\(\mathcal{P} = \{p_i\}_{i=1}^n\), where each \(p_i \in \mathbb{R}^3\)
anchors a physically meaningful spatial constraint in the environment.

Through this process, the OOD input pair \((o,l)_{\textit{OOD}}\) is deterministically compressed into
a geometric scaffold \(\mathcal{P}\),
which exposes the spatial variables required for downstream, differentiable reward evaluation.

\subsubsection{Programmatic Reward Generation}

Given the grounded keypoint set \(\mathcal{P}\),
VLS leverages the cross-modal reasoning capability of VLMs to synthesize
stage-aware, programmatic reward functions.
Specifically, the VLM is queried to:
(i) decompose the task implied by \((o,l)_{\textit{OOD}}\) into \(S\) sequential stages,
and (ii) for each stage \(s \in \{1,\dots,S\}\), generate a differentiable reward function
that evaluates how well an action proposal satisfies the corresponding spatial constraints.

Formally, for each stage \(s\), the VLM produces a reward function $\mathcal{R}_s(\action_{t:t+T}^k,(o,l)_{\textit{OOD}})
\;=\;
f_{\text{VLM}}(\action_{t:t+T}^k,\mathcal{P},s),$
where \(\mathcal{R}_s\) defines a stage-specific potential field over the action space,
returning a scalar value that is differentiable with respect to the action proposal
\(\action_{t:t+T}^k\).
Intuitively, \(\mathcal{R}_s\) measures the degree to which the proposed action trajectory
respects the spatial relationships encoded by \(\mathcal{P}\) at stage \(s\).
To ensure differentiability, we constrain the VLM to output programmatic reward definitions
implemented as PyTorch~\cite{paszke2019pytorch} functions composed of differentiable tensor operations
(e.g., distances, dot products, soft constraints).
During inference, gradients are backpropagated through the instantiated reward function,
while the VLM itself remains a non-differentiable, off-graph component.

This construction directly instantiates the gradient guidance signal required for inference-time steering:
\begin{equation}\label{eq:g=R}
g_s
\;\triangleq\;
\nabla_{\action_{t:t+T}^k}
\mathcal{R}_s(\action_{t:t+T}^k,(o,l)_{\textit{OOD}}),
\end{equation}
providing a dense, action-space gradient that approximates
\(\nabla_{\action_{t:t+T}^k} \log p((o,l)_{\textit{OOD}} \mid \action_{t:t+T}^k)\).

\subsection{Action Denoising Process Guidance}
\label{subsection 4.2}

Given the stage-wise reward functions \(\{\mathcal{R}_s\}\) constructed in Sec.~\ref{subsection 4.1},
we next describe how VLS injects their gradients into the denoising process of the frozen base policy \(\pi^\star\).
Our goal is to steer inference-time sampling toward action trajectories that satisfy the constraints implied by the OOD input pair \((o,l)_{\textit{OOD}}\),
while preserving diversity and robustness under complex, multi-modal landscapes.
To this end, we combine gradient-based refinement with particle-level diversity and gradient-free resampling.

\subsubsection{Diverse Proposal Initialization with Repulsive Forces}

At each environment timestep \(t\), denoising begins by sampling a batch of \(B\) action proposals
\(\{\action_{t:t+T}^K[i] \sim \mathcal{N}(\mathbf{0},\mathbf{I})\}_{i=1}^B\) independently.
To prevent the batch from collapsing prematurely to a narrow mode of the base policy,
we introduce a diversity-promoting repulsive force during the early denoising steps.
Inspired by~\cite{tdp,particleguidance}, we define a repulsive gradient based on pairwise distances:
\begin{equation}
g_{\textit{RBF}}^k[i]
\;=\;
\nabla_{\action_{t:t+T}^k[i]}
\sum_{j \neq i}
\frac{1}{\|\action_{t:t+T}^k[i] - \action_{t:t+T}^k[j]\|_2 + \epsilon}.
\end{equation}
This term encourages action proposals within the batch to remain separated,
ensuring broad coverage of the action manifold and providing diverse candidates
for subsequent reward-based guidance.

\subsubsection{Gradient-Based Refinement}

To bias the denoising trajectory toward actions that satisfy the OOD-induced constraints,
we inject the stage-specific reward gradient
\(g_s = \nabla_{\action_{t:t+T}^k}\mathcal{R}_s(\action_{t:t+T}^k,(o,l)_{\textit{OOD}})\)
into the denoising updates.
Specifically, following Eq.~(\ref{eq:diffusion_guidance}) for diffusion policies
and Eq.~(\ref{eq:flow_guidance}) for flow-matching policies,
the reward gradient is added to the noise or velocity prediction at each denoising step \(k\),
steering samples toward regions of higher reward. To improve stability under noisy gradients,
we adopt stochastic refinement with multiple inner updates per denoising step,
analogous to MCMC-based guidance~\cite{du2025dynaguide,wang2024itps,DuMCMC}.
This procedure enables smoother exploration of the reward landscape
and mitigates sensitivity to local gradient artifacts.

\subsubsection{Gradient-Free Resampling via Feynman--Kac Steering}

In addition to gradient-based refinement, VLS employs a gradient-free resampling mechanism
based on Feynman--Kac (FK) steering~\cite{delmoral2004feynman,doucet2001sequential,singhal2025general}.
We interpret the batch of action proposals as an interacting particle system
and periodically resample particles according to reward-based potentials.
For the \(i\)-th particle at denoising step \(k\), the potential is defined as
\begin{equation}\label{eq:potential_defination}
G_i^k
\;=\;
\exp\!\big(\mathcal{R}_s(\action_{t:t+T}^k[i],(o,l)_{\textit{OOD}})\big).
\end{equation}
Normalized weights \(w_i^k = G_i^k / \sum_{j=1}^B G_j^k\) are computed,
and multinomial resampling is applied to the particle set.
This procedure effectively tilts the transition kernel of the generative process
toward the target distribution
\(p(\action \mid (o,l)_{\textit{OOD}})\),
allowing high-reward particles to replicate while pruning proposals
that violate the OOD-induced constraints.

By combining continuous gradient-based steering with discrete, reward-weighted resampling,
VLS enables the frozen base policy \(\pi^\star\) to navigate complex,
multi-modal constraint landscapes efficiently,
avoiding the sample inefficiency and brittleness of purely selection-based methods.

\subsection{Closed-Loop Execution Control and Stage Switching}
\label{subsection 4.3}

To handle physical uncertainty (e.g., object slippage, partial execution)
and to robustly coordinate multi-stage tasks,
VLS incorporates a closed-loop execution control mechanism.
This mechanism uses execution feedback to
(i) adaptively regulate the guidance strength \(\lambda\) at the level of action chunks,
and (ii) determine when to switch between stage-specific reward functions
\(\{\mathcal{R}_s\}\) during task execution.

% \subsection{Closed-Loop VLM Verification}
% \label{subsection 4.3}

% To handle physical uncertainty (e.g., object slippage or partial execution)
% and to ensure stable transitions across task stages,
% VLS augments inference-time steering with a closed-loop control mechanism.
% This mechanism adaptively modulates the guidance strength \(\lambda\)
% and selectively queries a vision--language model (VLM) to monitor task progress
% under the OOD input pair \((o,l)_{\textit{OOD}}\).

\subsubsection{Adaptive Guidance Strength}

Within a fixed task stage \(s\),
multiple action chunks \(\{\action_{t:t+T}\}\) may be generated sequentially.
We adapt the guidance strength \(\lambda_t\) for each action chunk \(t\)
based on the relative reward achieved under the current stage-specific reward function \(\mathcal{R}_s\).

Let \(\mathcal{R}_s^t\) denote the reward value of the final denoising step
for the action chunk generated at chunk index \(t\) under stage \(s\),
and let \(\mathcal{R}_s^{\textit{base}}\) denote the corresponding reward
obtained from the first action chunk generated for this stage.
The guidance strength is computed as:
\begin{equation}\label{eq:adaptive_guidance}
\lambda_t
\;=\;
\lambda_{\max}
\cdot
\mathrm{sigmoid}
\!\left(
1 - \frac{\mathcal{R}_s^t}{\mathcal{R}_s^{\textit{base}}}
\right).
\end{equation}
This adaptive schedule increases guidance when the current action chunk
deviates from the constraints encoded by \(\mathcal{R}_s\),
and progressively reduces guidance as execution improves.
As a result, strong steering is applied when coarse correction is required,
while the frozen base policy \(\pi^\star\) is allowed to dominate
during fine-grained manipulation near stage completion.

% At environment timestep \(t\), we adapt the guidance strength \(\lambda_t\)
% based on the relative reward achieved by the current action chunk.
% Let \(\mathcal{R}_t\) denote the reward value of the final denoising step
% for the chunk generated at timestep \(t\),
% and let \(\mathcal{R}_{\textit{base}}\) denote the corresponding reward
% obtained from the first chunk generated for the current stage.
% We compute the guidance strength via a sigmoid mapping:
% \begin{equation}\label{eq:adaptive_guidance}
% \lambda_t
% \;=\;
% \lambda_{\max}
% \cdot
% \mathrm{sigmoid}
% \!\left(
% 1 - \frac{\mathcal{R}_t}{\mathcal{R}_{\textit{base}}}
% \right).
% \end{equation}
% This schedule increases the influence of reward-based steering
% when the current action violates the OOD-induced constraints,
% and gradually reduces guidance as the reward improves.
% As a result, steering dominates when coarse correction is required,
% while the base policy \(\pi^\star\) is allowed to handle fine-grained manipulation
% near successful execution.

\subsubsection{Schmitt-Trigger-Based Stage Switching}

To robustly determine when to transition between stages
and to avoid oscillatory behavior near stage boundaries,
we employ a hysteresis-based switching mechanism inspired by the Schmitt trigger~\cite{schmitt1938thermionic}.
For the current stage \(s\), we define two reward thresholds
\(R_{\textit{high}}\) and \(R_{\textit{low}}\),
and compute a switching signal \(Q_t\) based on the evolution of \(\mathcal{R}_s^t\):

\begin{equation}\label{eq:Schmitt}
Q_t =
\begin{cases}
\text{Advance stage}, & \mathcal{R}_s^t > R_{\textit{high}}, \\
\text{Maintain stage}, & R_{\textit{low}} \le \mathcal{R}_s^t \le R_{\textit{high}}, \\
\text{Reinforce stage}, & \mathcal{R}_s^t < R_{\textit{low}} .
\end{cases}
\end{equation}

When a switching event is triggered,
a vision--language model is queried to interpret the execution outcome
and select the appropriate next-stage reward function \(\mathcal{R}_{s+1}\),
or to continue applying the current stage reward \(\mathcal{R}_s\)
with updated guidance strength.
By introducing hysteresis into stage switching,
VLS avoids premature transitions and repeated oscillations,
enabling stable coordination across stages
under physical uncertainty and complex OOD execution dynamics. The whole algorithm of VLS can be found at Algorithm \ref{alg:vls}.

% To avoid excessive VLM queries and to prevent oscillatory behavior near stage boundaries,
% we employ a hysteresis-based verification mechanism inspired by the Schmitt trigger~\cite{schmitt1938thermionic}.
% Specifically, we define a query trigger \(Q_t\) using two reward thresholds,
% \(R_{\text{high}}\) and \(R_{\text{low}}\):
% \begin{equation}\label{eq:Schmitt}
% Q_t
% =
% \begin{cases}
% \text{Verify-Success}, & \text{if } \mathcal{R}_t \text{ crosses } R_{\text{high}} \text{ from below}, \\
% \text{Check-Failure}, & \text{if } \mathcal{R}_t \text{ crosses } R_{\text{low}} \text{ from above}, \\
% \text{Idle}, & \text{otherwise}.
% \end{cases}
% \end{equation}
% When \(Q_t\) is triggered, the VLM is queried to assess the current task stage
% and determine whether the system should advance to the next stage,
% maintain the current stage with continued steering,
% or re-engage guidance due to execution failure.
% By combining hysteresis-based verification with adaptive guidance strength,
% VLS mitigates the brittleness of open-loop generative execution
% and enables robust multi-stage task execution
% under physical uncertainty and complex OOD inputs.

\begin{algorithm}[t]
\small
\SetCommentSty{\scriptsize\ttfamily}
\caption{VLS Algorithm}
\label{alg:vls}

\KwIn{
Base policy $\pi^\star$;
Initial observation $o_{0}$;
language instruction $l$;
chunk horizon $T$;
sample batch size $B$
}
\KwOut{Action chunk $a_{t:t+T}$}

\texttt{// Condition grounding and reward generation}\;
$\mathcal{P} = \{p_i\}_{i=1}^n \leftarrow LVM(o_{0}, l)$

$\{\mathcal{R}_s(\action_{t:t+T}, \mathcal{P})\}_{s=1}^S \leftarrow f_{\textit{VLM}}(o_{0}, l, \mathcal{P})$

\texttt{// Initialize parameters}\;
$s \leftarrow 1$\;
$MCMC \leftarrow 4$ if $\pi^\star$ is diffusion else $1$\;

\texttt{// Denoising loop at action chunk index $t$}\;
Sample initial proposals:
$\{\action_{t:t+T}^K[i] \sim \mathcal{N}(0,I)\}_{i=1}^B$\;

\For{$k = K \rightarrow 0$}{
    \texttt{// Diversity initialization}\;
    $g_{\textit{RBF}}^k[i] =
    \nabla_{\action_{t:t+T}^k[i]}
    \sum_{j \neq i}
    \frac{1}{\|\action_{t:t+T}^k[i] - \action_{t:t+T}^k[j]\|_2 + \epsilon}$

    Use $g_{\textit{RBF}}^k$ as $g$ in Eq.~(\ref{eq:diffusion_guidance}) or Eq.~(\ref{eq:flow_guidance})

    \texttt{// Gradient-based refinement}\;
    $g_{\textit{reward}}^k =
    \nabla_{\action_{t:t+T}^k}
    \mathcal{R}_s(\action_{t:t+T}^k, \mathcal{P})$\;

    \For{$m = 1 \rightarrow MCMC$}{
        Use $g_{\textit{reward}}^k$ as $g$ in Eq.~(\ref{eq:diffusion_guidance}) or Eq.~(\ref{eq:flow_guidance})
    }

    \texttt{// Gradient-free resampling}\;
    \For{$i = 1 \rightarrow B$}{
        $G_i^k \leftarrow \exp(\mathcal{R}_s(\action_{t:t+T}^k[i], \mathcal{P}))$\;
        $w_i^k \leftarrow G_i^k / \sum_{j=1}^B G_j^k$\;
    }
    Resample $\{\action_{t:t+T}^k[i]\}_{i=1}^B$ according to $\{w_i^k\}$
}

\texttt{// Closed-loop execution control}\;
Adapt $\lambda_t$ via Eq.~(\ref{eq:adaptive_guidance})

Update stage $s$ via Eq.~(\ref{eq:Schmitt})

\Return{$\action_{t:t+T}[0]$}
\end{algorithm}

\section{Experiments}

We evaluate (VLS) in both simulation and real-world settings. Simulation experiments are conducted on two widely used manipulation benchmarks, CALVIN~\cite{calvin} and LIBERO-PRO~\cite{zhou2025liberopro}, while real-world deployment is performed on a Franka Emika robot. We systematically test generalization under both spatial and semantic shifts. To explicitly model inference-time out-of-distribution (OOD) conditions, we introduce controlled perturbations at test time along two axes:

\textbf{Observation perturbations.}
We modify the environment state by (i) adding previously unseen objects as distractors, (ii) changing objects' attribute during testing, and (iii) changing the positions or orientations of task-relevant objects and support surfaces. 

\textbf{Language perturbations.}
We alter task instructions by changing target objects and goal behaviors. More details on the perturbation and task description are provided in  Appendix.

\begin{table*}[t]
\centering
\small
\setlength{\tabcolsep}{4pt}
\begin{tabular}{lccccccccccc}
\toprule
\multirow{2}{*}{\textbf{Method}}
& \multicolumn{5}{c}{\textbf{Task Perturbation}}
& \multicolumn{5}{c}{\textbf{Position Perturbation}}
& \multicolumn{1}{c}{\textbf{Overall}} \\
\cmidrule(lr){2-6}\cmidrule(lr){7-11}\cmidrule(lr){12-12}
& Goal & Spatial & 10 & Object & Avg.
& Goal & Spatial & 10 & Object & Avg.
& Avg. \\
\midrule
OpenVLA
 & 0.00 & 0.00 & 0.00 & 0.00 & 0.00
 & 0.00 & 0.00 & 0.00 & 0.00 & 0.00
 & 0.00 \\
$\pi$-0
 & 0.00 & 0.00 & 0.00 & 0.00 & 0.00
 & 0.00 & 0.00 & 0.00 & 0.00 & 0.00
 & 0.00 \\
$\pi$-05
 & 0.00 & 1.00 & 1.00 & 1.00 & 0.75
 & 38.00 & 20.00 & 8.00 & 17.00 & 20.75
 & 10.75 \\
$\pi$-0.5 (LeRobot)
 & 12.00 & 48.50 & 21.50 & 10.50 & 23.13
 & 29.00 & 41.00 & 11.00 & 16.00 & 24.25
 & 23.69 \\
$\pi$-0.5 (LeRobot) + VLS
 & \textbf{33.50} & \textbf{54.00} & \textbf{25.50} & \textbf{41.00} & \textbf{38.50}
 & \textbf{38.00} & \textbf{42.00} & \textbf{15.50} & \textbf{45.00} & \textbf{35.13}
 & \textbf{36.81} \\
\bottomrule
\end{tabular}

\caption{\textbf{LIBERO-PRO results.} We test VLA baselines and a frozen $\pi_{0.5}$ policy with/without VLS. The experimental environment consists of LIBERO-PRO~\cite{zhou2025liberopro}'s task and position perturbation, applied to LIBERO~\cite{liu2023libero}'s four suites: Goal, Spatial, 10 (Long) and Object, with each suite containing 10 tasks. For each task in each suite, we test 20 episodes and report the average success rates (\%). ``Overall'' reports the mean across all columns.}
\label{table:table1}
\end{table*}

Our evaluation answers the following questions:

\begin{itemize}
    \item \textbf{Q1. Is inference-time steering necessary to handle observation and language shifts at test time?}
    \item \textbf{Q2. Does VLS provide stronger adaptation than existing inference-time steering approaches?}
    \item \textbf{Q3. What is the contribution of each component in the VLS framework?}
    \item \textbf{Q4. Can VLS adapt policies in the real world with minimal computational overhead?}
\end{itemize}

\begin{figure*}[t]
    \centering
    \includegraphics[width=\textwidth]{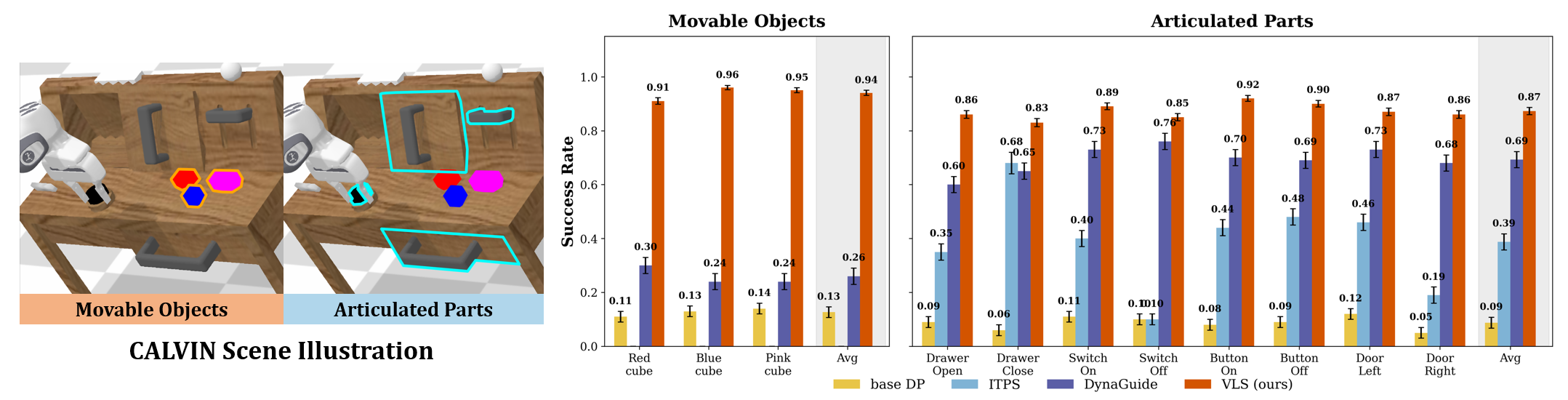}
    \caption{\textbf{Steering methods comparison on CALVIN.} Success rates for VLS (ours), DynaGuide, ITPS, and the base diffusion policy across movable objects (cubes) and articulated parts (drawer, switch, button, door). VLS achieves 94\% average on movable objects (7.4$\times$ over base policy) and 87\% on articulated parts (9.6$\times$ boost), outperforming prior steering methods by 15--25 percentage points. Error bars show standard deviation over 600 episodes per task.}
    \label{fig:comparison_with_steering_methods}
\end{figure*}

\subsection{Baselines}

We compare VLS against seven baselines, grouped into VLA models and inference-time steering methods. (Detailed Implementation in Appendix.)

\textbf{VLA models}: To answer Q1, we evaluate four leading VLA models that rank highly on the LIBERO-PRO leaderboard: OpenVLA~\cite{kim24openvla}, $\pi_0$~\cite{black2024pi0}, $\pi_{0.5}$~\cite{black2025pi05}, and $\pi_{0.5}$ LeRobot finetuned version~\cite{lerobot_pi05}. All models use a VLM backbone to jointly reason over observations and language instructions and are evaluated without any fine-tuning on our OOD test scenarios. 

\textbf{DP Steering policies}: To answer Q2, we compare against two popular inference-time steering approaches on the same frozen base policy: i) DynaGuide~\cite{du2025dynaguide}, which steers denoising using distances in pretrained DINO feature space as heuristic guidance; and ii) ITPS~\cite{wang2024itps}, which selects from a predefined set of guidance functions based on the detected OOD condition. 

% Both baselines rely on manually designed or discrete guidance mechanisms and do not condition guidance on a rich multimodal goal specification.

\textbf{Ablation}: To answer Q3, we evaluate three ablated variants of VLS that remove one component at a time: i) w/o gradient guidance, ii) w/o Feynman–Kac (FK) resampling, and iii) w/o RBF-based diversity initialization.

\begin{figure*}[t]
    \centering
    \includegraphics[width=\textwidth]{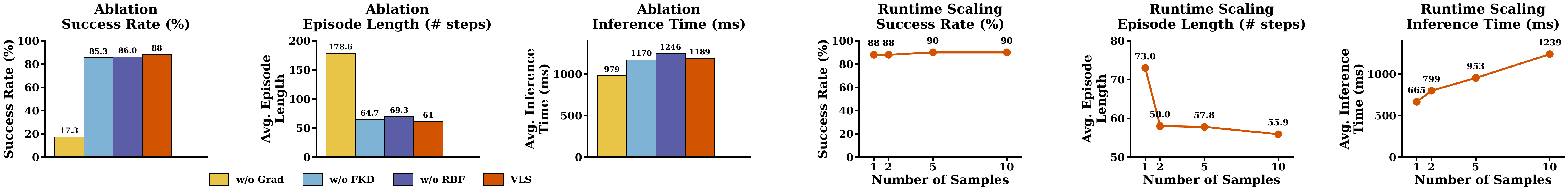}
    \caption{(left) Ablation of VLS components (50 episodes per task). We compare Full VLS (gradient guidance + FK steering + RBF diversity, with $K=10$) against variants that remove FK steering (w/o FKD), remove RBF diversity (w/o RBF), or remove gradient guidance (w/o grad). (right) Scaling with sample batch size $K$ on \texttt{door\_left} (50 episodes). Larger $K$ improves performance but increases inference time, illustrating a compute--performance tradeoff.}
    \label{fig:ablation_components}
\end{figure*}

\subsection{Results}

\textbf{Inference-Time Steering Is Necessary.} We choose LIBERO-PRO~\cite{zhou2025liberopro}, a simulation benchmark, as our testing platform. This is an OOD test suite developed based on LIBERO~\cite{liu2023libero}, which primarily includes comprehensive perturbations across five aspects of the original LIBERO's four task suites: object, position, semantic, task, and environment. Among these, the \textbf{position} and \textbf{task perturbations} best align with the description of OOD scenarios in this paper. The position perturbation refers to relocating objects ($o_{\textit{OOD}}$) while keeping language instructions unchanged. The task perturbation refers to redefining task logic and target states, where visual observations remain in-distribution while language instructions are completely changed ($l_{\textit{OOD}}$). For each perturbation for tasks in each suites, we text 20 episodes. We choose Success Rate (SR) as metric. We evaluate four leading VLA models that rank highly on the LIBERO-PRO leaderboard: OpenVLA~\cite{kim24openvla}, $\pi_0$~\cite{black2024pi0}, $\pi_{0.5}$~\cite{black2025pi05}, and $\pi_{0.5}$ LeRobot finetuned version~\cite{lerobot_pi05}. 

As shown in Table \ref{table:table1},
these pre-trained VLAs, despite leveraging pretrained VLM backbones for perception and instruction understanding, struggle to adapt to joint observation and language shifts at test time. VLS consistently outperforms all evaluated VLA models under joint observation and language perturbations (Table \ref{table:table1}). While VLAs exhibit strong in-distribution performance, their success rates drop sharply under OOD conditions. This failure persists despite the use of pretrained VLM backbones. We attribute this to the fact that post-training on robot data entangles spatial reasoning with specific training contexts, effectively degrading the VLM’s generalization ability when the execution environment deviates from the training manifold. These results shows inference-time steering is necessary for pretrained policy adaptation.

% preamble: \usepackage{booktabs}

\begin{figure*}[t]
    \centering
    \includegraphics[width=\textwidth]{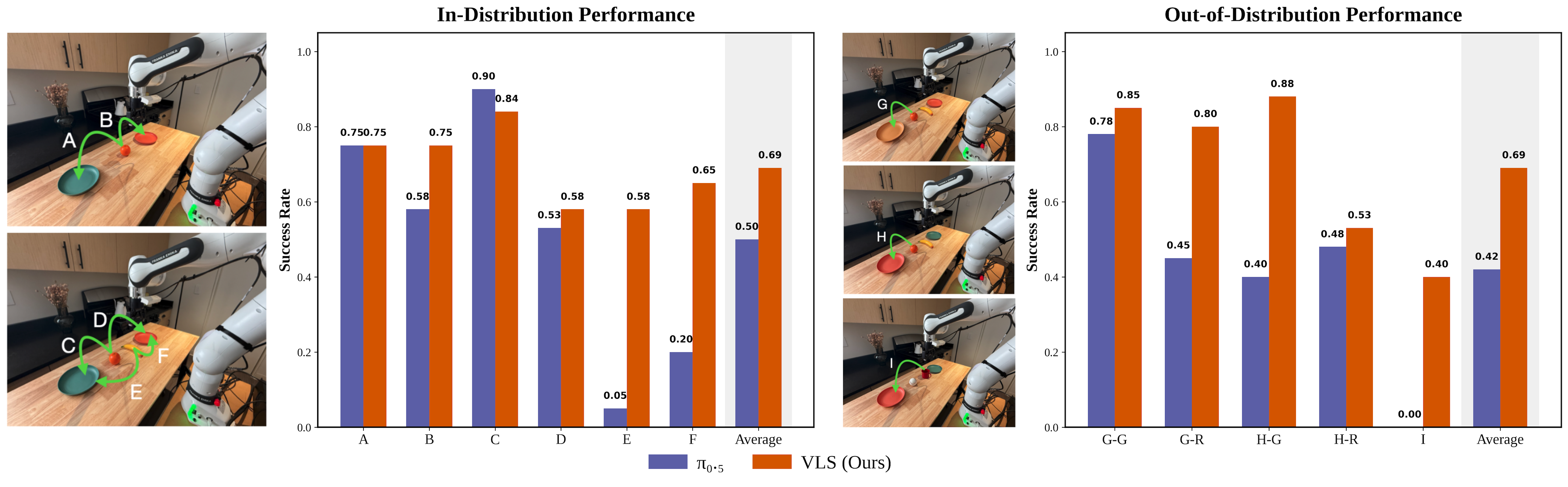}
    \caption{\textbf{Real-world Deployment on a Franka robot.} (\textbf{Left: In-distribution tasks}) Task layouts, language instructions, and success rates for in-distribution real-world manipulation.
Level~1 (top) requires placing an orange onto a specified plate (red or green) based on the instruction.
Level~2 (bottom) introduces an additional object (banana), requiring sequential selection of both the target object and the target plate.
Bar plots report per-task and average success rates for the frozen \(\pi\)-0.5 baseline and VLS. (\textbf{Right: Out-of-distribution tasks}) Task layouts, instructions, and results under test-time distribution shifts.
We evaluate three OOD variants:
(1) \emph{Appearance shift} (top), replacing the red/green plate with a previously unseen yellow plate;
(2) \emph{Position shift} (middle), swapping the locations of the two plates while keeping the instruction unchanged;
(3) \emph{Object shift} (bottom), replacing the banana with a never-before-seen mug and instructing the robot to place the mug on the green plate.
Each task is evaluated over 20 trials.
Grasping the correct object contributes 50\% success, and full task completion contributes 100\%.
VLS consistently outperforms the baseline and maintains robust execution under real-world OOD conditions.}
    \label{fig:real_world_deployment}
\end{figure*}

\textbf{VLS Outperforms Existing Steering Methods.} We compared with two leading steering methods, DynaGuide \cite{du2025dynaguide} and ITPS \cite{wang2024itps} on CALVIN~\cite{calvin}. As shown in left part of Figure \ref{fig:comparison_with_steering_methods}, where a Franka Panda robot interacts with a tabletop scene containing articulated objects (door, drawer, button, switch) and three randomly placed colored cubes (red, blue, pink). 

% The robot receives RGB observations from static and gripper-mounted cameras plus proprioception, and executes 7-DoF delta actions at 30\,Hz. A task succeeds when the target object state changes beyond a threshold (e.g., 60\% of joint range for articulated objects, 5\,mm displacement for cubes). 

As illustrated in Figure~\ref{fig:comparison_with_steering_methods}, all steering methods improve over the unsteered base diffusion policy, confirming the necessity of inference-time steering. On \emph{MovableObjects} (cube manipulation), VLS achieves a 94\% average success rate, corresponding to a 7.4$\times$ improvement over the base policy. On \emph{ArticulatedParts} (drawer, switch, button, door), VLS reaches 87\% average success, a 9.6$\times$ gain. ITPS performs reasonably on articulated tasks with fixed target states, but fails on movable-object tasks where object positions vary across episodes. DynaGuide improves performance across both task groups, but its DINO-feature-based heuristic lacks the expressiveness to capture task-specific spatial requirements. In contrast, VLS conditions its guidance directly on the current observation–language input, enabling precise steering under spatial variability and task-dependent constraints that heuristic guidance cannot reliably handle.

\textbf{Ablation Study of components.} We evaluate three ablated variants of VLS that remove one component at a time: i) w/o gradient guidance, ii) w/o Feynman–Kac (FK) resampling, and iii) w/o RBF-based diversity initialization.

\textit{Effect of gradient-based guidance.}
Removing gradient guidance causes a severe performance collapse across all tasks, with success rates dropping to near-failure and episode lengths increasing substantially (Figure \ref{fig:ablation_components} (left)). This confirms that dense, trajectory-differentiable guidance is the primary driver of VLS’s effectiveness.

\textit{Role of FK resampling and RBF diversity.}
Removing FK resampling or RBF-based diversity has a smaller impact on success rate but consistently degrades efficiency and stability. These components improve sample efficiency by preventing premature collapse to suboptimal modes and by maintaining global coverage early in denoising.

\textit{Scaling with sample batch size.}
As shown in Figure \ref{fig:ablation_components} (right), increasing the batch size improves success rates and reduces episode length, at the cost of higher inference latency. This exposes a practical compute–performance trade-off that can be tuned for deployment. 

Together, these results show that both gradient-free exploration and gradient-based refinement are necessary for robust inference-time control. Robust inference-time adaptation requires both gradient-free global exploration (to avoid poor initial modes) and gradient-based local refinement (to satisfy fine-grained constraints during execution).

\textbf{VLS Enables Efficient Real-World Deployment.} 
We evaluate VLS on a Franka Emika robot to test whether inference-time steering can reliably adapt a frozen VLA policy under real-world test-time variation. (Detailed Implementation in Appendix.)

As shown in Figure \ref{fig:real_world_deployment}, VLS consistently improves real-world task success over the frozen \(\pi\)-0.5 baseline across both in-distribution and out-of-distribution settings.
In in-distribution tasks requiring object selection and placement, VLS achieves a 69\% average success rate, outperforming the baseline by 19\%.
Under out-of-distribution conditions involving appearance changes, object repositioning, and novel object substitutions, the baseline performance degrades sharply, while VLS maintains stable execution and substantially higher success rates.
In the most challenging object-level OOD case, where the target object is replaced by a previously unseen mug, the baseline fails entirely, whereas VLS succeeds in 40\% of trials. These results show that VLS can be deployed efficiently in real robotic systems and enables pretrained policies to adapt to test-time spatial and semantic variation through inference-time steering alone.

\section{Conclusion \& Limitation}

We propose VLS, a training-free framework that guides pre-trained robotic policies using differentiable rewards generated by Vision–Language Models, addressing the challenge of policy deployment in OOD scenarios. Experiments demonstrate that VLS significantly outperforms existing methods in both simulation and real-world tasks. Limitations of VLS include computational latency: batch sampling, MCMC runs, and FK resampling introduce high inference overhead. Future work may explore progress-aware reward signal generation and optimizing computational efficiency during inference.

\balance
\bibliographystyle{plainnat}
\bibliography{references}

\begin{thebibliography}{55}
\providecommand{\natexlab}[1]{#1}
\providecommand{\url}[1]{\texttt{#1}}
\expandafter\ifx\csname urlstyle\endcsname\relax
  \providecommand{\doi}[1]{doi: #1}\else
  \providecommand{\doi}{doi: \begingroup \urlstyle{rm}\Url}\fi

\bibitem[Ahn et~al.(2022)Ahn, Brohan, Brown, Chebotar, Cortes, David, Finn, Fu, Gopalakrishnan, Hausman, et~al.]{ahn2022saycan}
Michael Ahn, Anthony Brohan, Noah Brown, Yevgen Chebotar, Omar Cortes, Byron David, Chelsea Finn, Chuyuan Fu, Keerthana Gopalakrishnan, Karol Hausman, et~al.
\newblock Do as i can, not as i say: Grounding language in robotic affordances.
\newblock \emph{arXiv preprint arXiv:2204.01691}, 2022.

\bibitem[Barreiros et~al.(2025)Barreiros, Beaulieu, Bhat, Cory, Cousineau, Dai, Fang, Hashimoto, Irshad, Itkina, et~al.]{barreiros2025careful}
Jose Barreiros, Andrew Beaulieu, Aditya Bhat, Rick Cory, Eric Cousineau, Hongkai Dai, Ching-Hsin Fang, Kunimatsu Hashimoto, Muhammad~Zubair Irshad, Masha Itkina, et~al.
\newblock A careful examination of large behavior models for multitask dexterous manipulation.
\newblock \emph{arXiv preprint arXiv:2507.05331}, 2025.

\bibitem[Black et~al.(2024)Black, Brown, Driess, Esmail, Equi, Finn, Fusai, Groom, Hausman, Ichter, Jakubczak, Jones, Ke, Levine, Li-Bell, Mothukuri, Nair, Pertsch, Shi, Tanner, Vuong, Walling, Wang, and Zhilinsky]{black2024pi0}
Kevin Black, Noah Brown, Danny Driess, Adnan Esmail, Michael Equi, Chelsea Finn, Niccolo Fusai, Lachy Groom, Karol Hausman, Brian Ichter, Szymon Jakubczak, Tim Jones, Liyiming Ke, Sergey Levine, Adrian Li-Bell, Mohith Mothukuri, Suraj Nair, Karl Pertsch, Lucy~Xiaoyang Shi, James Tanner, Quan Vuong, Anna Walling, Haohuan Wang, and Ury Zhilinsky.
\newblock $\pi_0$: A vision-language-action flow model for general robot control.
\newblock \emph{arXiv preprint arXiv:2410.24164}, 2024.

\bibitem[Black et~al.(2025)Black, Brown, Darpinian, Dhabalia, Driess, Esmail, Equi, Finn, Fusai, Galliker, Ghosh, Groom, Hausman, ichter, Jakubczak, Jones, Ke, LeBlanc, Levine, Li-Bell, Mothukuri, Nair, Pertsch, Ren, Shi, Smith, Springenberg, Stachowicz, Tanner, Vuong, Walke, Walling, Wang, Yu, and Zhilinsky]{black2025pi05}
Kevin Black, Noah Brown, James Darpinian, Karan Dhabalia, Danny Driess, Adnan Esmail, Michael~Robert Equi, Chelsea Finn, Niccolo Fusai, Manuel~Y. Galliker, Dibya Ghosh, Lachy Groom, Karol Hausman, brian ichter, Szymon Jakubczak, Tim Jones, Liyiming Ke, Devin LeBlanc, Sergey Levine, Adrian Li-Bell, Mohith Mothukuri, Suraj Nair, Karl Pertsch, Allen~Z. Ren, Lucy~Xiaoyang Shi, Laura Smith, Jost~Tobias Springenberg, Kyle Stachowicz, James Tanner, Quan Vuong, Homer Walke, Anna Walling, Haohuan Wang, Lili Yu, and Ury Zhilinsky.
\newblock $\pi_{0.5}$: a vision-language-action model with open-world generalization.
\newblock In Joseph Lim, Shuran Song, and Hae-Won Park, editors, \emph{Proceedings of The 9th Conference on Robot Learning}, volume 305 of \emph{Proceedings of Machine Learning Research}, pages 17--40. PMLR, 27--30 Sep 2025.

\bibitem[Brohan et~al.(2022)Brohan, Brown, Carbajal, Chebotar, Dabis, Finn, Gopalakrishnan, Hausman, Herzog, Hsu, Ibarz, Irpan, Jackson, Jesmonth, Joshi, Julian, Kalashnikov, Kuang, Leal, Lee, Levine, Lu, Malla, Manjunath, Mordatch, Nachum, Parada, Peralta, Pertsch, Quiambao, Rao, Ryoo, Salazar, Sanketi, Sayed, Singh, Sontakke, Stone, Tan, Tran, Vanhoucke, Vega, Vuong, Xia, Xiao, Xu, Xu, Yu, and Zitkovich]{rt12022arxiv}
Anthony Brohan, Noah Brown, Justice Carbajal, Yevgen Chebotar, Joseph Dabis, Chelsea Finn, Keerthana Gopalakrishnan, Karol Hausman, Alexander Herzog, Jasmine Hsu, Julian Ibarz, Alex Irpan, Tomas Jackson, Sally Jesmonth, Nikhil Joshi, Ryan Julian, Dmitry Kalashnikov, Yuheng Kuang, Isabel Leal, Kuang-Huei Lee, Sergey Levine, Yao Lu, Utsav Malla, Deeksha Manjunath, Igor Mordatch, Ofir Nachum, Carolina Parada, Jodilyn Peralta, Karl Pertsch, Jornell Quiambao, Kanishka Rao, Michael Ryoo, Grecia Salazar, Pannag Sanketi, Kevin Sayed, Jaspiar Singh, Sumedh Sontakke, Austin Stone, Clayton Tan, Huong Tran, Vincent Vanhoucke, Steve Vega, Quan Vuong, Fei Xia, Ted Xiao, Peng Xu, Sichun Xu, Tianhe Yu, and Brianna Zitkovich.
\newblock Rt-1: Robotics transformer for real-world control at scale.
\newblock In \emph{arXiv preprint arXiv:2212.06817}, 2022.

\bibitem[Brohan et~al.(2023)Brohan, Brown, Carbajal, Chebotar, Chen, Choromanski, Ding, Driess, Dubey, Finn, Florence, Fu, Gonzalez~Arenas, Gopalakrishnan, Han, Hausman, Herzog, Hsu, Ichter, Irpan, Joshi, Julian, Kalashnikov, Kuang, Leal, Lee, Lee, Levine, Lu, Michalewski, Mordatch, Pertsch, Rao, Reymann, Ryoo, Salazar, Sanketi, Sermanet, Singh, Singh, Soricut, Tran, Vanhoucke, Vuong, Wahid, Welker, Wohlhart, Wu, Xia, Xiao, Xu, Xu, Yu, and Zitkovich]{brohan2023rt2}
Anthony Brohan, Noah Brown, Justice Carbajal, Yevgen Chebotar, Xi~Chen, Krzysztof Choromanski, Tianli Ding, Danny Driess, Avinava Dubey, Chelsea Finn, Pete Florence, Chuyuan Fu, Montserrat Gonzalez~Arenas, Keerthana Gopalakrishnan, Kehang Han, Karol Hausman, Alexander Herzog, Jasmine Hsu, Brian Ichter, Alex Irpan, Nikhil Joshi, Ryan Julian, Dmitry Kalashnikov, Yuheng Kuang, Isabel Leal, Lisa Lee, Tsang-Wei~Edward Lee, Sergey Levine, Yao Lu, Henryk Michalewski, Igor Mordatch, Karl Pertsch, Kanishka Rao, Krista Reymann, Michael Ryoo, Grecia Salazar, Pannag Sanketi, Pierre Sermanet, Jaspiar Singh, Anikait Singh, Radu Soricut, Huong Tran, Vincent Vanhoucke, Quan Vuong, Ayzaan Wahid, Stefan Welker, Paul Wohlhart, Jialin Wu, Fei Xia, Ted Xiao, Peng Xu, Sichun Xu, Tianhe Yu, and Brianna Zitkovich.
\newblock Rt-2: Vision-language-action models transfer web knowledge to robotic control.
\newblock \emph{arXiv preprint arXiv:2307.15818}, 2023.

\bibitem[Cao et~al.(2025)Cao, Huang, Guo, Zhang, Nan, Mai, Wang, Cheng, Sun, Han, Zhao, Zhang, Guo, Zheng, Song, Li, Luo, and Luo]{yu2025compose}
Jiahang Cao, Yize Huang, Hanzhong Guo, Rui Zhang, Mu~Nan, Weijian Mai, Jiaxu Wang, Hao Cheng, Jingkai Sun, Gang Han, Wen Zhao, Qiang Zhang, Yijie Guo, Qihao Zheng, Chunfeng Song, Xiao Li, Ping Luo, and Andrew~F. Luo.
\newblock Compose your policies! improving diffusion-based or flow-based robot policies via test-time distribution-level composition.
\newblock \emph{arXiv preprint arXiv:2510.01068}, 2025.

\bibitem[Caron et~al.(2021)Caron, Touvron, Misra, J\'egou, Mairal, Bojanowski, and Joulin]{dino}
Mathilde Caron, Hugo Touvron, Ishan Misra, Herv\'e J\'egou, Julien Mairal, Piotr Bojanowski, and Armand Joulin.
\newblock Emerging properties in self-supervised vision transformers.
\newblock In \emph{Proceedings of the IEEE/CVF International Conference on Computer Vision (ICCV)}, pages 9650--9660, October 2021.

\bibitem[Chi et~al.(2023)Chi, Xu, Feng, Cousineau, Du, Burchfiel, Tedrake, and Song]{chi2023diffusionpolicy}
Cheng Chi, Zhenjia Xu, Siyuan Feng, Eric Cousineau, Yilun Du, Benjamin Burchfiel, Russ Tedrake, and Shuran Song.
\newblock Diffusion policy: Visuomotor policy learning via action diffusion.
\newblock \emph{arXiv preprint arXiv:2303.04137}, 2023.

\bibitem[Corso et~al.(2023)Corso, Xu, Bortoli, Barzilay, and Jaakkola]{particleguidance}
Gabriele Corso, Yilun Xu, Valentin~De Bortoli, Regina Barzilay, and T.~Jaakkola.
\newblock Particle guidance: non-i.i.d. diverse sampling with diffusion models.
\newblock \emph{ArXiv}, abs/2310.13102, 2023.
\newblock URL \url{https://api.semanticscholar.org/CorpusID:264405842}.

\bibitem[Dathathri et~al.(2019)Dathathri, Madotto, Lan, Hung, Frank, Molino, Yosinski, and Liu]{dathathri2019pplm}
Sumanth Dathathri, Andrea Madotto, Janice Lan, Jane Hung, Eric Frank, Piero Molino, Jason Yosinski, and Rosanne Liu.
\newblock Plug and play language models: A simple approach to controlled text generation.
\newblock \emph{arXiv preprint arXiv:1912.02164}, 2019.

\bibitem[Del~Moral(2004)]{delmoral2004feynman}
Pierre Del~Moral.
\newblock \emph{Feynman-Kac Formulae: Genealogical and Interacting Particle Systems with Applications}.
\newblock Springer, 2004.

\bibitem[Dhariwal and Nichol(2021)]{dhariwal2021guided}
Prafulla Dhariwal and Alex Nichol.
\newblock Diffusion models beat gans on image synthesis.
\newblock \emph{arXiv preprint arXiv:2105.05233}, 2021.

\bibitem[Doucet et~al.(2001)Doucet, de~Freitas, and Gordon]{doucet2001sequential}
Arnaud Doucet, Nando de~Freitas, and Neil Gordon.
\newblock \emph{Sequential Monte Carlo Methods in Practice}.
\newblock Springer, 2001.

\bibitem[Driess et~al.(2023)Driess, Xia, Sax, Ichter, Gopalakrishnan, Bohg, Zeng, Finn, Levine, Hausman, et~al.]{driess2023palme}
Danny Driess, Fei Xia, Alexander Sax, Brian Ichter, Keerthana Gopalakrishnan, Jeannette Bohg, Andy Zeng, Chelsea Finn, Sergey Levine, Karol Hausman, et~al.
\newblock {PaLM-E}: An embodied multimodal language model.
\newblock \emph{arXiv preprint arXiv:2303.03378}, 2023.

\bibitem[Du and Song(2025)]{du2025dynaguide}
Maximilian Du and Shuran Song.
\newblock Dynaguide: Steering diffusion polices with active dynamic guidance.
\newblock \emph{arXiv preprint arXiv:2506.13922}, 2025.

\bibitem[Du et~al.(2023)Du, Durkan, Strudel, Tenenbaum, Dieleman, Fergus, Sohl-Dickstein, Doucet, and Grathwohl]{DuMCMC}
Yilun Du, Conor Durkan, Robin Strudel, Joshua~B. Tenenbaum, Sander Dieleman, Rob Fergus, Jascha~Narain Sohl-Dickstein, A.~Doucet, and Will Grathwohl.
\newblock Reduce, reuse, recycle: Compositional generation with energy-based diffusion models and mcmc.
\newblock \emph{ArXiv}, abs/2302.11552, 2023.
\newblock URL \url{https://api.semanticscholar.org/CorpusID:257078922}.

\bibitem[Duan et~al.(2021)Duan, Jian, and Tan]{duan2021spacesimulatorphysicalinteractions}
Jiafei Duan, Samson Yu~Bai Jian, and Cheston Tan.
\newblock Space: A simulator for physical interactions and causal learning in 3d environments, 2021.
\newblock URL \url{https://arxiv.org/abs/2108.06180}.

\bibitem[Duan et~al.(2022)Duan, Dasgupta, Fischer, and Tan]{duan2022survey}
Jiafei Duan, Arijit Dasgupta, Jason Fischer, and Cheston Tan.
\newblock A survey on machine learning approaches for modelling intuitive physics.
\newblock \emph{arXiv preprint arXiv:2202.06481}, 2022.

\bibitem[Ho and Salimans(2022)]{ho2022cfg}
Jonathan Ho and Tim Salimans.
\newblock Classifier-free diffusion guidance.
\newblock \emph{arXiv preprint arXiv:2207.12598}, 2022.

\bibitem[Ho et~al.(2020)Ho, Jain, and Abbeel]{ho_ddpm_2020}
Jonathan Ho, Ajay Jain, and Pieter Abbeel.
\newblock {DDPM}: Denoising diffusion probabilistic models.
\newblock In \emph{Advances in Neural Information Processing Systems}, volume~33, pages 6840--6851, 2020.

\bibitem[Huang et~al.(2023)Huang, Wang, Zhang, Li, Wu, and Fei-Fei]{huang2023voxposer}
Wenlong Huang, Chen Wang, Ruohan Zhang, Yunzhu Li, Jiajun Wu, and Li~Fei-Fei.
\newblock Voxposer: Composable 3d value maps for robotic manipulation with language models.
\newblock \emph{arXiv preprint arXiv:2307.05973}, 2023.

\bibitem[Huang et~al.(2024)Huang, Wang, Li, Zhang, and Fei-Fei]{huang2024rekep}
Wenlong Huang, Chen Wang, Yunzhu Li, Ruohan Zhang, and Li~Fei-Fei.
\newblock Rekep: Spatio-temporal reasoning of relational keypoint constraints for robotic manipulation.
\newblock \emph{arXiv preprint arXiv:2409.01652}, 2024.

\bibitem[Jeon et~al.(2025)Jeon, Min, and Park]{tdp}
Hyeonseong Jeon, Cheolhong Min, and Jaesik Park.
\newblock Tree-guided diffusion planner.
\newblock 2025.
\newblock URL \url{https://api.semanticscholar.org/CorpusID:280985003}.

\bibitem[Jiang et~al.(2022)Jiang, Gupta, Zhang, Wang, Dou, Chen, Fei-Fei, Anandkumar, Zhu, and Fan]{jiang2022vima}
Yunfan Jiang, Agrim Gupta, Zichen Zhang, Guanzhi Wang, Yongqiang Dou, Yanjun Chen, Li~Fei-Fei, Anima Anandkumar, Yuke Zhu, and Linxi Fan.
\newblock Vima: General robot manipulation with multimodal prompts.
\newblock \emph{arXiv preprint arXiv:2210.03094}, 2022.

\bibitem[Khazatsky et~al.(2024)Khazatsky, Pertsch, Nair, Balakrishna, Dasari, Karamcheti, Nasiriany, Srirama, Chen, Ellis, et~al.]{khazatsky2024droid}
Alexander Khazatsky, Karl Pertsch, Suraj Nair, Ashwin Balakrishna, Sudeep Dasari, Siddharth Karamcheti, Soroush Nasiriany, Mohan~Kumar Srirama, Lawrence~Yunliang Chen, Kirsty Ellis, et~al.
\newblock Droid: A large-scale in-the-wild robot manipulation dataset.
\newblock \emph{arXiv preprint arXiv:2403.12945}, 2024.

\bibitem[Kim et~al.(2022)Kim, Kwon, and Ye]{kim2022diffusionclip}
Gwanghyun Kim, Taesung Kwon, and Jong~Chul Ye.
\newblock Diffusionclip: Text-guided diffusion models for robust image manipulation.
\newblock \emph{arXiv preprint arXiv:2110.02711}, 2022.

\bibitem[Kim et~al.(2024)Kim, Pertsch, Karamcheti, Xiao, Balakrishna, Nair, Rafailov, Foster, Lam, Sanketi, Vuong, Kollar, Burchfiel, Tedrake, Sadigh, Levine, Liang, and Finn]{kim24openvla}
{Moo Jin} Kim, Karl Pertsch, Siddharth Karamcheti, Ted Xiao, Ashwin Balakrishna, Suraj Nair, Rafael Rafailov, Ethan Foster, Grace Lam, Pannag Sanketi, Quan Vuong, Thomas Kollar, Benjamin Burchfiel, Russ Tedrake, Dorsa Sadigh, Sergey Levine, Percy Liang, and Chelsea Finn.
\newblock Openvla: An open-source vision-language-action model.
\newblock \emph{arXiv preprint arXiv:2406.09246}, 2024.

\bibitem[Kirillov et~al.(2023)Kirillov, Mintun, Ravi, Mao, Rolland, Gustafson, Xiao, Whitehead, Berg, Lo, Dollar, and Girshick]{sam}
Alexander Kirillov, Eric Mintun, Nikhila Ravi, Hanzi Mao, Chloe Rolland, Laura Gustafson, Tete Xiao, Spencer Whitehead, Alexander~C. Berg, Wan-Yen Lo, Piotr Dollar, and Ross Girshick.
\newblock Segment anything.
\newblock In \emph{Proceedings of the IEEE/CVF International Conference on Computer Vision (ICCV)}, pages 4015--4026, October 2023.

\bibitem[Kumar et~al.(2024)Kumar, Shen, Ramos, Fox, Lozano-P{\'e}rez, Kaelbling, and Garrett]{kumar2024open}
Nishanth Kumar, William Shen, Fabio Ramos, Dieter Fox, Tom{\'a}s Lozano-P{\'e}rez, Leslie~Pack Kaelbling, and Caelan~Reed Garrett.
\newblock Open-world task and motion planning via vision-language model inferred constraints.
\newblock \emph{arXiv preprint arXiv:2411.08253}, 2024.

\bibitem[{LeRobot Team}(2025)]{lerobot_pi05}
{LeRobot Team}.
\newblock $\pi_{0.5}$ (pi05 libero) (lerobot).
\newblock \url{https://huggingface.co/lerobot/pi05_libero_finetuned}, 2025.
\newblock Model checkpoint + documentation (accessed 2026-01-31).

\bibitem[Li et~al.(2025)Li, Liu, Dong, Teng, Rouxel, Caldwell, and Chen]{li2025towards}
Zhuo Li, Junjia Liu, Zhipeng Dong, Tao Teng, Quentin Rouxel, Darwin Caldwell, and Fei Chen.
\newblock Towards deploying vla without fine-tuning: Plug-and-play inference-time vla policy steering via embodied evolutionary diffusion.
\newblock \emph{arXiv preprint arXiv:2511.14178}, 2025.

\bibitem[Lipman et~al.(2022)Lipman, Chen, Ben-Hamu, Nickel, and Le]{lipman2022flowmatching}
Yaron Lipman, Ricky T.~Q. Chen, Heli Ben-Hamu, Maximilian Nickel, and Matt Le.
\newblock Flow matching for generative modeling.
\newblock In \emph{arXiv preprint arXiv:2210.02747}, 2022.

\bibitem[Liu et~al.(2023)Liu, Zhu, Gao, Feng, Liu, Zhu, and Stone]{liu2023libero}
Bo~Liu, Yifeng Zhu, Chongkai Gao, Yihao Feng, Qiang Liu, Yuke Zhu, and Peter Stone.
\newblock Libero: Benchmarking knowledge transfer for lifelong robot learning.
\newblock \emph{arXiv preprint arXiv:2306.03310}, 2023.

\bibitem[Mees et~al.(2021)Mees, Hermann, Rosete-Beas, and Burgard]{calvin}
Oier Mees, Luk{\'a}s Hermann, Erick Rosete-Beas, and Wolfram Burgard.
\newblock Calvin: A benchmark for language-conditioned policy learning for long-horizon robot manipulation tasks.
\newblock \emph{IEEE Robotics and Automation Letters}, 7:\penalty0 7327--7334, 2021.
\newblock URL \url{https://api.semanticscholar.org/CorpusID:244908821}.

\bibitem[Meng et~al.(2021)Meng, He, Song, Song, Wu, Zhu, and Ermon]{meng2021sdedit}
Chenlin Meng, Yutong He, Yang Song, Jiaming Song, Jiajun Wu, Jun-Yan Zhu, and Stefano Ermon.
\newblock Sdedit: Guided image synthesis and editing with stochastic differential equations.
\newblock \emph{arXiv preprint arXiv:2108.01073}, 2021.

\bibitem[Nakamoto et~al.(2024)Nakamoto, Mees, Kumar, and Levine]{nakamoto2024vgps}
Mitsuhiko Nakamoto, Oier Mees, Aviral Kumar, and Sergey Levine.
\newblock Steering your generalists: Improving robotic foundation models via value guidance.
\newblock \emph{arXiv preprint arXiv:2410.13816}, 2024.

\bibitem[O’Neill et~al.(2024)O’Neill, Rehman, Maddukuri, Gupta, Padalkar, Lee, Pooley, Gupta, Mandlekar, Jain, et~al.]{o2024open}
Abby O’Neill, Abdul Rehman, Abhiram Maddukuri, Abhishek Gupta, Abhishek Padalkar, Abraham Lee, Acorn Pooley, Agrim Gupta, Ajay Mandlekar, Ajinkya Jain, et~al.
\newblock Open x-embodiment: Robotic learning datasets and rt-x models: Open x-embodiment collaboration 0.
\newblock In \emph{2024 IEEE International Conference on Robotics and Automation (ICRA)}, pages 6892--6903. IEEE, 2024.

\bibitem[Paszke et~al.(2019)Paszke, Gross, Massa, Lerer, Bradbury, Chanan, Killeen, Lin, Gimelshein, Antiga, et~al.]{paszke2019pytorch}
Adam Paszke, Sam Gross, Francisco Massa, Adam Lerer, James Bradbury, Gregory Chanan, Trevor Killeen, Zeming Lin, Natalia Gimelshein, Luca Antiga, et~al.
\newblock Pytorch: An imperative style, high-performance deep learning library.
\newblock \emph{Advances in neural information processing systems}, 32, 2019.

\bibitem[Pumacay et~al.(2024)Pumacay, Singh, Duan, Krishna, Thomason, and Fox]{pumacay2024colosseum}
Wilbert Pumacay, Ishika Singh, Jiafei Duan, Ranjay Krishna, Jesse Thomason, and Dieter Fox.
\newblock The colosseum: A benchmark for evaluating generalization for robotic manipulation.
\newblock \emph{arXiv preprint arXiv:2402.08191}, 2024.

\bibitem[Reed et~al.(2022)Reed, Zolna, Parisotto, Colmenarejo, Novikov, Barth-Maron, Gimenez, Sulsky, Kay, Springenberg, et~al.]{reed2022gato}
Scott Reed, Konrad Zolna, Emilio Parisotto, Sergio~G{\'o}mez Colmenarejo, Alexander Novikov, Gabriel Barth-Maron, Manon Gimenez, Yevgenii Sulsky, Jack Kay, Jost~Tobias Springenberg, et~al.
\newblock A generalist agent.
\newblock \emph{arXiv preprint arXiv:2205.06175}, 2022.

\bibitem[Ross et~al.(2011)Ross, Gordon, and Bagnell]{ross2011dagger}
St{\'e}phane Ross, Geoffrey~J. Gordon, and J.~Andrew Bagnell.
\newblock A reduction of imitation learning and structured prediction to no-regret online learning.
\newblock \emph{Proceedings of the Fourteenth International Conference on Artificial Intelligence and Statistics}, 2011.

\bibitem[Schmitt(1938)]{schmitt1938thermionic}
O.~H. Schmitt.
\newblock A thermionic trigger.
\newblock \emph{Journal of Scientific Instruments}, 15\penalty0 (1):\penalty0 24--26, 1938.

\bibitem[Singhal et~al.(2025)Singhal, Horvitz, Teehan, Ren, Yu, McKeown, and Ranganath]{singhal2025general}
Raghav Singhal, Zachary Horvitz, Ryan Teehan, Mengye Ren, Zhou Yu, Kathleen McKeown, and Rajesh Ranganath.
\newblock A general framework for inference-time scaling and steering of diffusion models.
\newblock \emph{arXiv preprint arXiv:2501.06848}, 2025.

\bibitem[Sun and Song(2025)]{sun2025lpb}
Zhanyi Sun and Shuran Song.
\newblock Latent policy barrier: Learning robust visuomotor policies by staying in-distribution.
\newblock \emph{arXiv preprint arXiv:2508.05941}, 2025.

\bibitem[Tumanyan et~al.(2023)Tumanyan, Geyer, Bagon, and Dekel]{tumanyan2023pnpdiffusionfeatures}
Narek Tumanyan, Michael Geyer, Shai Bagon, and Tali Dekel.
\newblock Plug-and-play diffusion features for text-driven image-to-image translation.
\newblock \emph{arXiv preprint arXiv:2211.12572}, 2023.

\bibitem[Wagenmaker et~al.(2025)Wagenmaker, Nakamoto, Zhang, Park, Yagoub, Nagabandi, Gupta, and Levine]{wagenmaker2025dsrl}
Andrew Wagenmaker, Mitsuhiko Nakamoto, Yunchu Zhang, Seohong Park, Waleed Yagoub, Anusha Nagabandi, Abhishek Gupta, and Sergey Levine.
\newblock Steering your diffusion policy with latent space reinforcement learning.
\newblock \emph{arXiv preprint arXiv:2506.15799}, 2025.

\bibitem[Walke et~al.(2023)Walke, Black, Zhao, Vuong, Zheng, Hansen-Estruch, He, Myers, Kim, Du, et~al.]{walke2023bridgedata}
Homer~Rich Walke, Kevin Black, Tony~Z Zhao, Quan Vuong, Chongyi Zheng, Philippe Hansen-Estruch, Andre~Wang He, Vivek Myers, Moo~Jin Kim, Max Du, et~al.
\newblock Bridgedata v2: A dataset for robot learning at scale.
\newblock In \emph{Conference on Robot Learning}, pages 1723--1736. PMLR, 2023.

\bibitem[Wang et~al.(2024)Wang, Wang, Du, Sundaralingam, Yang, Chao, Perez-D'Arpino, Fox, and Shah]{wang2024itps}
Yanwei Wang, Lirui Wang, Yilun Du, Balakumar Sundaralingam, Xuning Yang, Yu-Wei Chao, Claudia Perez-D'Arpino, Dieter Fox, and Julie Shah.
\newblock Inference-time policy steering through human interactions.
\newblock \emph{arXiv preprint arXiv:2411.16627}, 2024.

\bibitem[Wu et~al.(2025{\natexlab{a}})Wu, Li, Hermans, Ramos, Bajcsy, and P{\'e}rez-D'Arpino]{wu2025dowhatyousay}
Yilin Wu, Anqi Li, Tucker Hermans, Fabio Ramos, Andrea Bajcsy, and Claudia P{\'e}rez-D'Arpino.
\newblock Do what you say: Steering vision-language-action models via runtime reasoning-action alignment verification.
\newblock \emph{arXiv preprint arXiv:2510.16281}, 2025{\natexlab{a}}.

\bibitem[Wu et~al.(2025{\natexlab{b}})Wu, Tian, Swamy, and Bajcsy]{wu2025forewarn}
Yilin Wu, Ran Tian, Gokul Swamy, and Andrea Bajcsy.
\newblock From foresight to forethought: {VLM}-in-the-loop policy steering via latent alignment.
\newblock \emph{arXiv preprint arXiv:2502.01828}, 2025{\natexlab{b}}.

\bibitem[Ye(2025)]{ye2025vgd}
Hanming Ye.
\newblock Steering diffusion policies with value-guided denoising.
\newblock In \emph{OpenReview (Forum Paper)}, 2025.
\newblock URL \url{https://openreview.net/forum?id=dtMBW9W5jo}.

\bibitem[Yuan et~al.(2024)Yuan, Mu, Tao, Fang, Zhang, and Su]{yuan2024policydecorator}
Xiu Yuan, Tongzhou Mu, Stone Tao, Yunhao Fang, Mengke Zhang, and Hao Su.
\newblock Policy decorator: Model-agnostic online refinement for large policy model.
\newblock \emph{arXiv preprint arXiv:2412.13630}, 2024.

\bibitem[Zhou et~al.(2025)Zhou, Xu, Tie, Chen, Zhang, Chu, Zhou, and Sun]{zhou2025liberopro}
Xueyang Zhou, Yangming Xu, Guiyao Tie, Yongchao Chen, Guowen Zhang, Duanfeng Chu, Pan Zhou, and Lichao Sun.
\newblock {LIBERO-PRO}: Towards robust and fair evaluation of vision-language-action models beyond memorization.
\newblock \emph{arXiv preprint arXiv:2510.03827}, 2025.

\bibitem[Zhu et~al.(2025)Zhu, Li, Yuan, Zhang, Liu, Zhang, Yu, Zhang, and Liu]{zhu2025usr}
Zhengbang Zhu, Ziyan Li, Xiu Yuan, Hanbo Zhang, Yuxiao Liu, Chongjie Zhang, Yong Yu, Weinan Zhang, and Minghuan Liu.
\newblock Unified latent steering and residual refinement for online improvement of diffusion policy models.
\newblock In \emph{ICLR 2026 Conference Submission (OpenReview)}, 2025.
\newblock URL \url{https://openreview.net/forum?id=DbBD2aT1OG}.

\end{thebibliography}

% \clearpage
% \onecolumn
% \appendix
% \setcounter{subsection}{0}
% \renewcommand{\thesubsection}{\Alph{subsection}}
% \renewcommand{\theHsubsection}{app.\Alph{subsection}}

% \input{rss_steerdp/appendix/gradient_guidance_derivation}
% \input{rss_steerdp/appendix/VLS_implementation}
% \input{rss_steerdp/appendix/tasks}
% \input{rss_steerdp/appendix/baselines}

\end{document}